\documentclass[10pt,conference]{IEEEtran}
\IEEEoverridecommandlockouts
\usepackage{subfigure}
\usepackage{cite}
\usepackage{amsmath,amssymb,amsfonts}
\usepackage{algorithmic}
\usepackage{graphicx}
\usepackage{textcomp}
\usepackage{xcolor}
\usepackage{url}
\usepackage[most]{tcolorbox}
\usepackage{cleveref}
\usepackage{booktabs}
\usepackage{colortbl}
\usepackage{threeparttable}
\usepackage{multirow}
\usepackage{enumitem}
\def\BibTeX{{\rm B\kern-.05em{\sc i\kern-.025em b}\kern-.08em
    T\kern-.1667em\lower.7ex\hbox{E}\kern-.125emX}}
\begin{document}

\title{Root Cause Analysis for Microservice Systems via Cascaded Conditional Learning with Hypergraphs}

\author{
Shuaiyu Xie\textsuperscript{1,*}\thanks{* Shuaiyu Xie and Hanbin He contributed equally to this work.},
Hanbin He\textsuperscript{1,*}, 
Jian Wang\textsuperscript{1,2,†}\thanks{† Jian Wang and Bing Li are the corresponding authors.},
Bing Li\textsuperscript{1,2,†}
\\[3pt]
\textsuperscript{1}\textit{Wuhan University, Wuhan, China} \\
\textsuperscript{2}\textit{Zhongguancun Laboratory, Beijing, China} \\
Email: theory@whu.edu.cn, hhbsgsh@whu.edu.cn, jianwang@whu.edu.cn, bingli@whu.edu.cn
\\[3pt]
}

\maketitle

\begin{abstract}
Root cause analysis in microservice systems typically involves two core tasks: root cause localization (RCL) and failure type identification (FTI). Despite substantial research efforts, conventional diagnostic approaches still face two key challenges. First, these methods predominantly adopt a joint learning paradigm for RCL and FTI to exploit shared information and reduce training time. However, this simplistic integration neglects the causal dependencies between tasks, thereby impeding inter-task collaboration and information transfer. Second, these existing methods primarily focus on point-to-point relationships between instances, overlooking the group nature of inter-instance influences induced by deployment configurations and load balancing. To overcome these limitations, we propose CCLH, a novel root cause analysis framework that orchestrates diagnostic tasks based on cascaded conditional learning. CCLH provides a three-level taxonomy for group influences between instances and incorporates a heterogeneous hypergraph to model these relationships, facilitating the simulation of failure propagation. Extensive experiments conducted on datasets from three microservice benchmarks demonstrate that CCLH outperforms state-of-the-art methods in both RCL and FTI.
\end{abstract}


\section{Introduction}
Microservice architecture has been widely adopted by cloud-native enterprises due to its flexibility, scalability, and loose coupling. In microservice systems (MSS), each microservice typically reproduces multiple instances, which collaborate with instances affiliated with other microservices to handle user requests \cite{xie2024tvdiag, zhang2024failure}. As these systems scale up, they may suffer from reliability issues, aka failures, attributable to the increasing complexity and dynamicity. Worse still, diagnosing failures in microservice systems is labor-intensive and time-consuming, due to the intricate failure propagation and the overwhelming volume of telemetry data. For example, GitHub once took approximately one and a half hours to resolve a failure that disrupted the codespace service, affecting millions of developers and repositories \cite{github}.

Traditional root cause analysis (RCA) in MSS encompasses two tasks: root cause localization (RCL) and failure type identification (FTI). Site Reliability Engineers (SREs) need to dissect excessive telemetry data to pinpoint the culprit component (e.g., the codespace service) and infer the failure type (e.g., CPU overload). It remains a laborious and error-prone endeavor, even for veteran SREs. To alleviate the burden on SREs and achieve prompt recovery, AIOps researchers have explored various automated RCA methods, reducing human intervention in RCL \cite{lee2023eadro, pham2024baro, tao2024diagnosing} and FTI \cite{tao2024giving, sui2023logkg}. Furthermore, some recent works have sought to jointly optimize these two tasks by incorporating deep learning-based methods, aiming to capture shared representations and reduce repetitive processing efforts \cite{xie2024tvdiag, sun2024art, zhang2023robust}.

Although existing multitask learning techniques perform competitively in RCA, several challenges persist. \textbf{The most critical issue arises from the insufficient exploitation of group-level relationships between microservice instances}. In MSS, microservices typically spawn multiple instances deployed across distributed hosts to support scalability and fault tolerance. However, the distributed nature and the load balancing mechanism complicate the propagation of failures between instances. For example, we assume a scenario where an instance with a high CPU limit encounters CPU overhead. This can have two cascading effects. On the one hand, the performance degradation of this instance may affect all upstream instances that maintain synchronous communication with it. On the other hand, an excessively high CPU limit can preempt more CPU time, degrading the performance of other co-located instances on the shared host. Ultimately, the overwhelmed instance may incur a failure state. As a result, the load-balancing mechanism will route traffic to other instances affiliated with the same microservice, further exacerbating their resource shortages. Most existing methods focus exclusively on call-level influences \cite{lee2023eadro, zhang2024trace}. Moreover, they often oversimplify inter-instance influences as point-to-point relationships, overlooking complex group-level interactions such as resource contention among co-located instances.

\textbf{The second issue lies in the parallel learning mechanism adopted for the diagnostic tasks}. Current multitask diagnostic methods ignore the inherent causal order between diagnostic tasks like RCL and FTI, leading to a disconnect between the identified culprit components and their corresponding failure types. For example, recent practices such as DiagFusion \cite{zhang2023robust} and TVDiag \cite{xie2024tvdiag} simultaneously optimize RCL and FTI by performing a weighted sum of their losses in an objective function. However, optimizing diagnostic tasks in parallel deviates from the intrinsic intuition of SREs in real-world scenarios. Since a microservice failure can trigger broader system anomalies due to the ripple effect, SREs typically begin by localizing the culprit component to eliminate interference from the excessive failure noise. Only after identifying the culprit component do they perform a more focused analysis of its telemetry data to determine the specific failure type \cite{zhang2024survey}. Therefore, it is more practical to design a sequential diagnosis approach that first performs root cause localization, followed by failure type identification. Such a design better reflects real-world diagnostic workflows and improves the interpretability and reliability of failure diagnosis.

In addressing these concerns, we propose CCLH, a root cause analysis framework designed to pinpoint culprit components and further classify the types of their failures. CCLH consists of three stages: (1) \textit{Multimodal Feature Extraction}. Given heterogeneous multimodal telemetry data, CCLH extracts and fuses the temporal features of each modality using gated recurrent units (GRUs) and cross-modal attention mechanisms (\cref{sec:MFE}). (2) \textit{Hypergraph-based Status Fusion}. We provide a three-level taxonomy for the diverse relationships between instances. More specifically, we utilize a hypergraph to fuse all instance status through heterogeneous hyperedges, endowing CCLH with the elaborate ability to perceive group influences (\cref{sec:HSF}). (3) \textit{Cascaded Conditional Learning and Diagnosis}. Recognizing the inherent causal order between diagnostic tasks, CCLH devises a cascaded conditional learning mechanism to orchestrate tasks during both training and inference phases, facilitating inter-task collaboration and information transfer (\cref{sec:CCL}). The major contributions of CCLH can be summarized as follows:

\begin{itemize}
    \item We present CCLH, a root cause analysis framework that coordinates different diagnostic tasks via a cascaded conditional learning mechanism.
    \item We introduce a three-level taxonomy for the diverse relationships between instances and design a heterogeneous hypergraph to perceive group influences in RCA.
    \item Extensive experiments are conducted on a public dataset and two popular microservice systems. 
\end{itemize}

\section{Background}
\label{sec:background}
\subsection{Root Cause Analysis in MSS}
\textbf{Basic Concepts}. The microservice architecture decomposes traditional monolithic software into granular and loosely coupled service units. Although the paradigm shift indeed facilitates the development process, it inevitably introduces failures due to unprecedented complexity and dynamicity. \textit{Failures} are unexpected incidents characterized by performance degradation or even service outages within MSS \cite{pham2024baro}. To avoid financial loss and reputational damage, SREs must promptly diagnose failures and implement appropriate recovery strategies after detecting anomalies \cite{xie2024tvdiag, roy2024exploring}. The life cycle of root cause analysis (RCA) involves root cause localization (RCL) and failure type identification (FTI). The goal of RCL is to locate the culprit component, namely the originating instance responsible for the current failure. FTI follows as the subsequent step, where SREs infer the underlying failure reason (e.g., file descriptor exhaustion) of the identified component. Failures of the culprit component can propagate to other instances through sophisticated inter-instance relationships, such as synchronous calls, resource contention, and load balancing, rendering RCA a laborious and error-prone task, even for experienced SREs. This highlights the need for an automated diagnostic tool capable of accurately narrowing the range of candidate microservices and determining the corresponding failure types.

\textbf{Multimodal Telemetry Data}. After failures are detected, the foremost step is to acquire failure-related telemetry to prepare for the subsequent diagnosis. Telemetry data, which includes metrics, traces, and logs, serves as the cornerstone of observability in the MSS. Metrics capture the key performance indicators in the form of time series. Traces are tree-like structures that record the processing trajectory of requests and the execution status of microservices in MSS. Logs document system events with timestamps, messages, and recorders, typically in a semi-structured format. Although telemetry offers significant practical benefits, the sheer volume of data leaves SREs preoccupied with filtering and processing, thereby impeding their ability to rapidly consolidate failure-related information.

\textbf{Problem Formulation}. Given metrics $\mathcal{M}$, traces $\mathcal{T}$, and logs $\mathcal{L}$, we group and transform them into a set of features $H=\{h_v^\mathcal{M}, h_v^\mathcal{T}, h_v^\mathcal{L}|v\in\mathcal{V}\}$, where $\mathcal{V}$ denotes the set of instances and $\left(h_v^\mathcal{M}, h_v^\mathcal{T}, h_v^\mathcal{L}\right)$ denotes the processed features of metrics, traces, and logs from instance $v$. Subsequently, we aim to design an automated approach $\mathcal{D}$ for root cause analysis, which assigns scores to all instances, indicating their likelihood of being the culprit component. Furthermore, the approach $D$ anatomizes the predicted culprit component for failure type identification. The process can be formulated as $\mathcal{D}\left(H \right) \to t, \mathcal{S}$, where $\mathcal{S}=\{s_1,s_2,...,s_v|v\in\mathcal{V}\}$ denotes scores of all instances, and $t$ represents the failure type. 

\subsection{Hypergraph}
Typically, instances within MSS manifest group relationships rather than isolated pairwise patterns. For example, the co-located instances, which are deployed on the same host, often compete for shared resources, including CPU time, memory, and network bandwidth. An improper configuration of an instance may induce collective starvation, thereby affecting all co-located instances. Similarly, when a particular instance crashes, the resulting traffic is distributed across multiple remaining instances under the same microservice. Such group-affecting behaviors are difficult to capture solely using traditional graph structures that restrict connections as point-to-point relationships (i.e., A is connected to B).

\begin{figure*}[t]
\centering{
 \subfigure[ Call relationship ]{
 \includegraphics[width=0.30\textwidth]{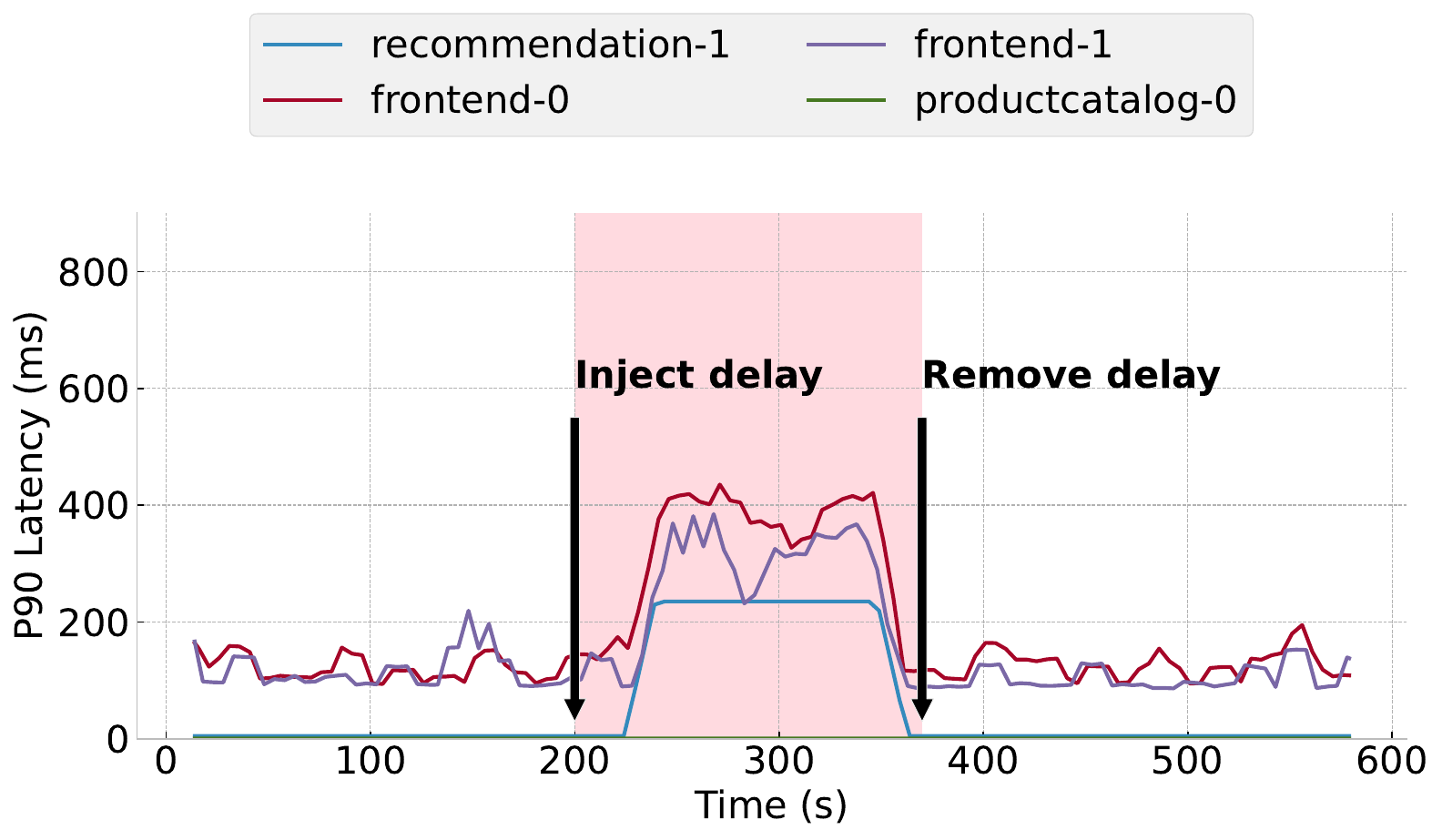} } 
 \subfigure[ Deployment relationship ]{  \includegraphics[width=0.32\textwidth]{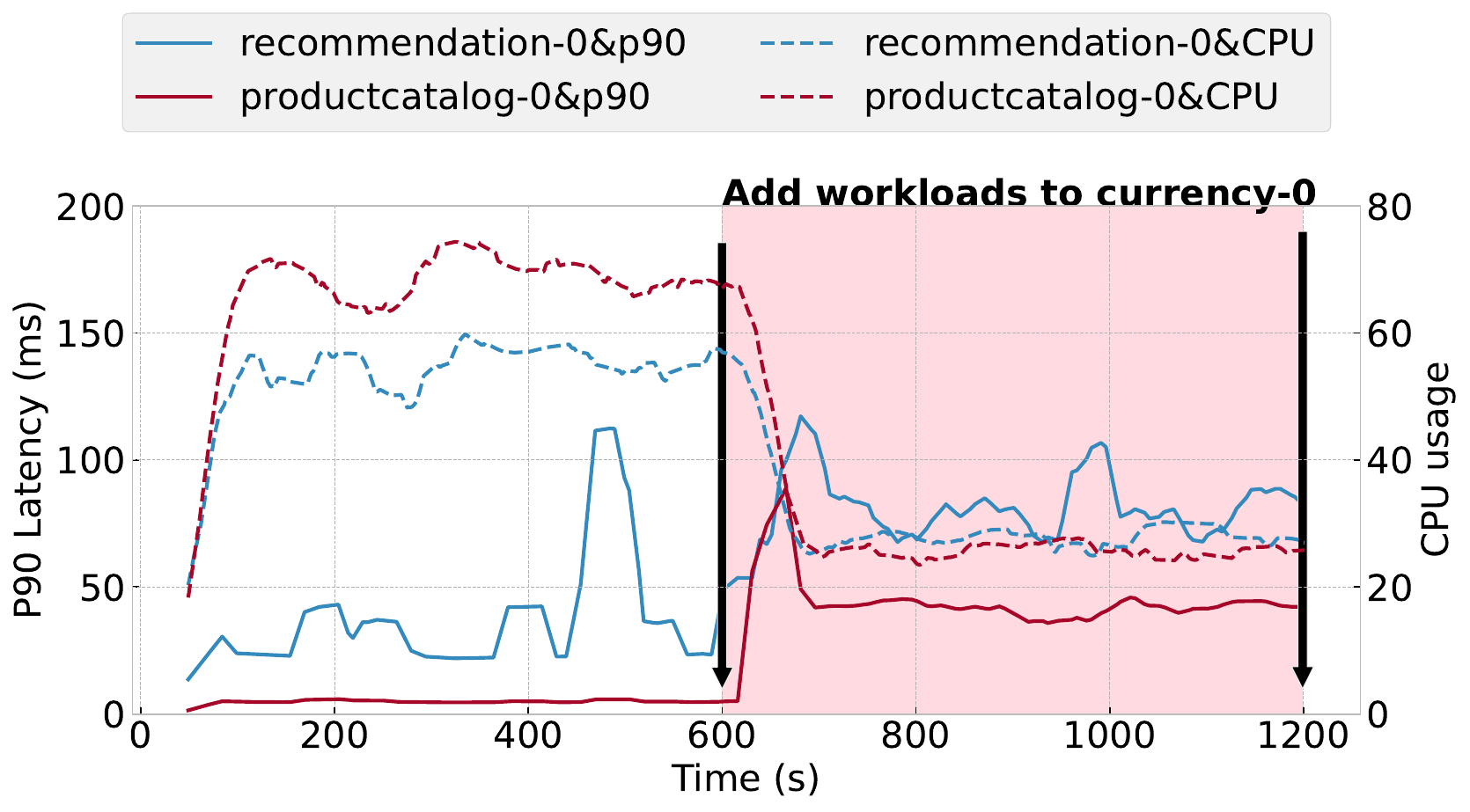}}
 \subfigure[ Load balancing relationship ]{  \includegraphics[width=0.32\textwidth]{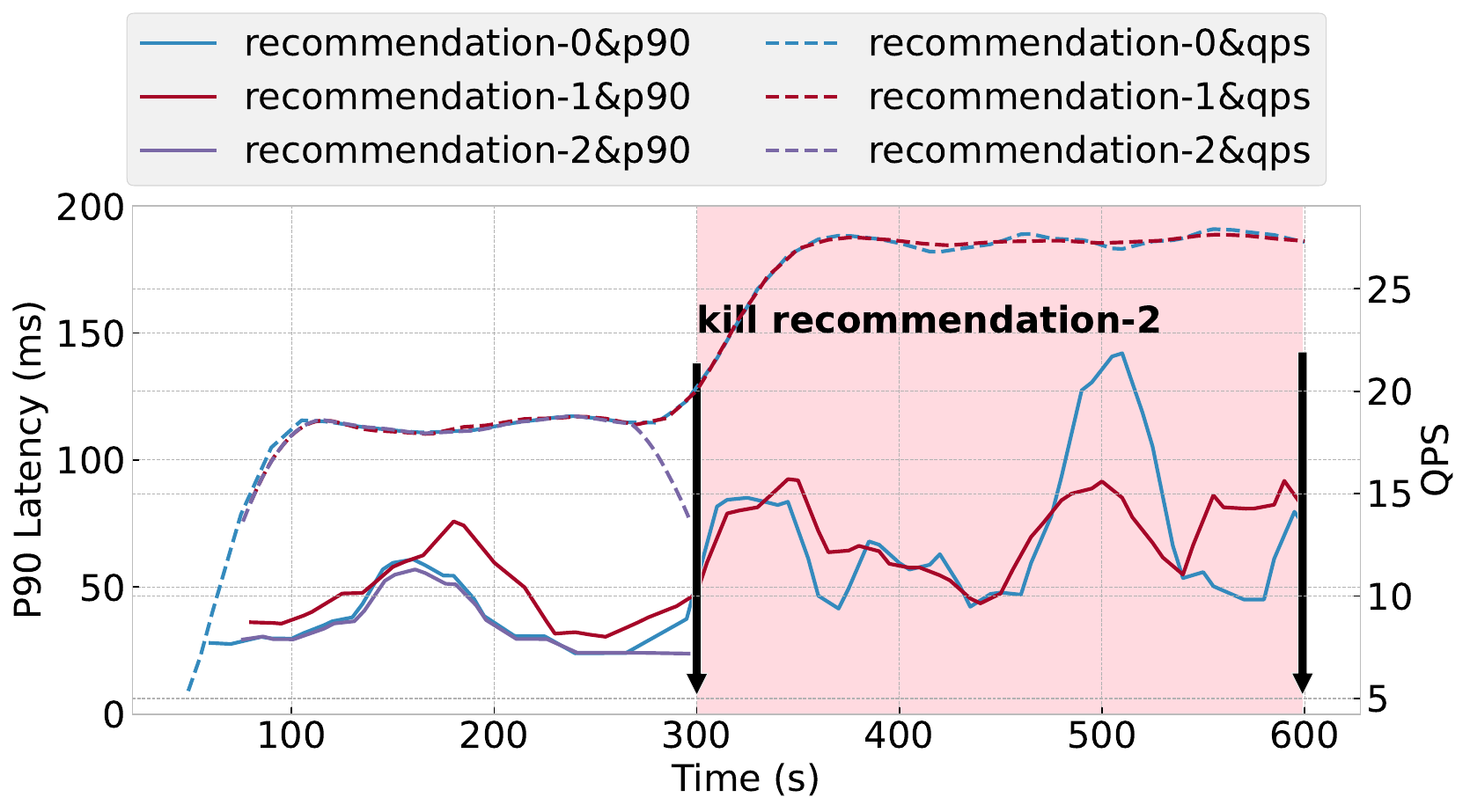}
}}
\caption{The group relationships between instances.}
\label{fig:collective-association}
\end{figure*}

\begin{figure}
    \centering
    \includegraphics[width=0.8\linewidth]{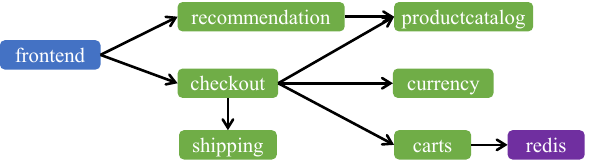}
    \caption{Part of call relationships in Online Boutique.}
    \label{fig:hipster}
\end{figure}

Hypergraphs, which allow each edge to link any number of vertices, inherently capture these group relationships. Formally, a hypergraph is defined as $HG=(V, E)$, where $V$ denotes the set of nodes, and $E$ is the set of hyperedges, each representing a subset of interacting nodes. This distinctive structure effectively represents the many-to-many relationships inherent in MSS, such as resource contention and load balancing. To handle the complexities and expressiveness of hypergraphs, many studies \cite{bai2021hypergraph, feng2019hypergraph, ijcai21-UniGNN, arya2020hypersage} have made initial endeavors in the development of hypergraph neural networks (HGNNs), a deep learning architecture tailored to hypergraph learning. Due to the superiority of HGNNs in hypergraph learning, CCLH implements HGNNs based on a popular framework, namely UniGAT \cite{ijcai21-UniGNN}, and further incorporates heterogeneous hyperedges to individualize different group relationships between instances.

\section{Observations and Insights}
In this section, we investigate two key aspects of root cause analysis: the group relationships between instances and the causal relationships between diagnostic tasks in root cause analysis. Our observations are based on experiments conducted using a widely adopted microservice system,  OnlineBoutique \cite{hipster}. Fig. \ref{fig:hipster} shows the partial call relationships in the system. We deploy Online Boutique in a local cluster and initialize several instances for each microservice. For example, ``frontend-0" and ``frontend-1" are two subordinate instances of the ``frontend" microservice.

\subsection{Group relationships between instances}
\label{sec:motivation2}
In traditional monolithic software, failures propagate between functions in a linear and point-to-point manner. In contrast, in MSS, failure propagation among instances often manifests an explosive pattern, where multiple instances, both directly and indirectly associated, are affected simultaneously. This is largely due to the diverse group relationships that exist among instances in MSS. In this section, we summarize three typical group relationships and their corresponding failure patterns, including call, deployment, and load balancing. 

\textbf{Call}. As shown in Fig. \ref{fig:collective-association} (a), we inject the network delay into the instance ``recommendation-1",  which results in a significant increase in latency, thus degrading the performance of all upstream callers (``frontend-0" and ``frontend-1"). In contrast, the downstream instance ``productcatalog-0" remains unaffected, maintaining stable latency due to the directional nature of the call relationship.

\textbf{Deployment}. Co-located instances deployed on the same host are logically independent but compete for shared resources such as CPU. We deploy three instances, including ``recommendation-0", ``productcatalog-0", and ``currency-0", on a resource-limited host. Note that ``currency-0" has no direct call relationships with the other two instances. To examine the impact of resource contention, we send requests to all instances while progressively increasing the traffic load on ``currency-0". We also lift the resource limits on ``currency-0" to simulate failures caused by unrestrained resource allocation. Fig. \ref{fig:collective-association} (b) illustrates the fluctuation of host resources and instance performance. As CPU usage of ``currency-0" increases, ``recommendation-0" and ``productcatalog-0'' incur a significant performance degradation due to the round-robin CPU scheduling mechanism. This is further evidenced by the sustained decline in CPU usage of ``recommendation-0" and ``productcatalog-0". The experiment confirms that resource contention can affect all co-located instances, regardless of logical dependencies, as they share the same physical host.

\textbf{Load Balancing}. We simulate an eviction scenario caused by insufficient machine resources by deliberately killing an instance  (``recommendation-2"). As depicted in Fig. \ref{fig:collective-association} (c), once ``recommendation-2" is terminated, the system’s load balancing mechanism redirects its unprocessed requests to sibling instances (i.e., ``recommendation-0" and ``recommendation-1"). This results in an increased workload for these instances, leading to a significant rise in their response latency. This type of influence is inherently group-based, as all sibling instances are collectively impacted by the failure of one. The impact becomes particularly severe when the traffic is heavy and the number of available sibling instances is limited.

	\begin{tcolorbox}[
		colback=white,
		colframe=gray,
		width=\linewidth,
		arc=1.5mm, auto outer arc,
		left = 1mm, 
		right = 1mm,
		top = 1mm,
		bottom = 1mm,
		breakable]		
		\textbf{Insight 1:} Unlike the linear and point-to-point failure propagation typically observed in monolithic software systems, inter-instance influences in MSS are more diverse and exhibit group-level effects. These effects are primarily driven by factors such as shared resource contention and load balancing mechanisms. This observation highlights the need for more accurate modeling techniques, such as hypergraphs, capable of capturing complex group relationships beyond simple pairwise relations.
	\end{tcolorbox}

\begin{figure*}[t]
    \centering
\includegraphics[width=\textwidth]{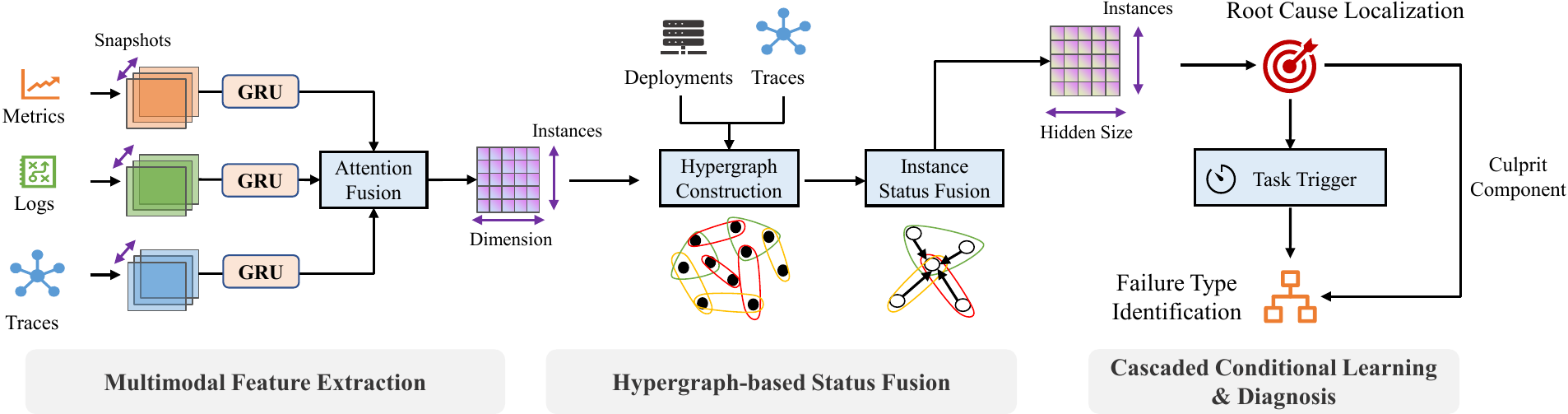}
    \caption{Overall architecture of CCLH.}
    \label{fig:structure}
\end{figure*}

\subsection{Causal order across diagnostic tasks}
\label{sec:motivation1}
Most existing approaches \cite{zhang2023robust, xie2024tvdiag, sun2024art, lee2023eadro} jointly train diagnostic tasks (RCL and FTI) using a shared objective or model. This multitask learning mechanism improves the performance of relevant tasks by focusing on shared knowledge and reducing overfitting \cite{fifty2021efficiently, zhang2021survey}. However, these methods overlook the intrinsic dependency between two diagnostic tasks, i.e., the outcome of one task can meaningfully inform and facilitate the other. As a result, performing RCL and FTI in parallel can lead to inconsistencies or contradictions between the identified culprit components and associated failure types.

Take Fig. \ref{fig:collective-association} (c) as an example. If SREs attempt to determine the failure type directly without first performing RCL, they may intuitively attribute the issue to CPU overhead based on the observed latency in traces and the CPU utilization of the two surviving instances in metrics. In contrast, if the culprit component (e.g., ``recommendation-2") is localized beforehand, the diagnostic process can be focused specifically on its associated telemetry data, enabling faster and more accurate FTI.

Unfortunately, existing methods rely on a parallel training paradigm that prevents FTI from harnessing the narrowed search space identified by RCL. The FTI model must ingest features from all instances indiscriminately based on techniques like pooling, which introduces noise and hampers generalizability, especially in scenarios with numerous and dynamic instances. This insight is further evidenced by the experimental results depicted in \cref{sec:RQ1}. Consequently, we argue that diagnostic tasks should follow a sequential paradigm, where RCL is executed first to constrain the scope for FTI. This principle is embedded in the design of CCLH, which aligns with practical diagnostic workflows and allows FTI to focus on a subset of likely faulty components.

\begin{tcolorbox}[
		colback=white,
		colframe=gray,
		width=\linewidth,
		arc=1.5mm, auto outer arc,
		left = 1mm, 
		right = 1mm,
		top = 1mm,
		bottom = 1mm,
		breakable]		
		\textbf{Insight 2:} Existing multitask diagnostic frameworks adopt parallel training and inference strategies, overlooking the inherent causal dependencies between diagnostic tasks. This limitation impedes effective coordination and information flow across tasks, ultimately constraining diagnostic accuracy and interpretability. 
	\end{tcolorbox}

\section{Methodology}

\subsection{Overview}
Inspired by the preceding insights, we propose a novel root cause analysis approach, namely CCLH, which cascades the localization of culprit components and the identification of their corresponding failure types. Fig. \ref{fig:structure} illustrates the overall architecture of CCLH. CCLH is composed of three phases: (1) \textit{Multimodal Feature Extraction}. To unify the representation of multimodal heterogeneous telemetry, CCLH uses gated recurrent units and a cross-modal attention mechanism to extract temporal features at the instance level. (2) \textit{Hypergraph-based Status Fusion}. Given the features of all instances, CCLH constructs a hypergraph to integrate their status. More specifically, we design heterogeneous hyperedges based on the taxonomy of group relationships among instances, enabling CCLH to tailor representations for various relationships. (3) \textit{Cascaded Conditional Learning and Diagnosis}. In this phase, CCLH cascades diagnostic tasks in the multitask training phase, emphasizing the inherent inter-task causality and preserving the shared knowledge. Specifically, we leverage instance-level features to predict the culprit component and then infer its failure type. Considering that the predicted culprit component is the prerequisite for FTI, we devise a two-stage training strategy: the RCL task is first trained independently to ensure reliable performance, after which the FTI task is introduced in the subsequent stage.

During the online diagnosis phase, CCLH still follows the same inference order as in the training phase, performing RCL first to pinpoint the culprit component, followed by FTI on the predicted instance.

\subsection{Multimodal Feature Extraction}
\label{sec:MFE}
\subsubsection{Data Preprocessing}
After MSS undergoes failures, the first step typically taken by SREs is to collect sufficient telemetry data during the failure period \cite{roy2024exploring}. As discussed in \cref{sec:background}, multimodal telemetry data is heterogeneous and voluminous. Therefore, it is advisable to transform this telemetry into unified time series representations that facilitate subsequent analysis and learning. 

Given a sliding window with both the length and the stride set to $\tau$ (30s by default), CCLH uniformly segments the failure period into a sequence of snapshots, where each snapshot represents the state of the MSS within the corresponding time window. Subsequently, we serialize the multimodal telemetry for each instance. For metrics, due to their temporal nature, we compute the average value of each metric within every snapshot. For traces, we calculate the average instance duration and count the occurrences of each status code within a snapshot. For logs, CCLH first parses raw logs using a widely adopted log parser, namely Drain \cite{he2017drain}, to extract fixed templates. Then, we count the frequency of each template within each snapshot. After preprocessing, we concatenate all snapshots to construct a multimodal sequence set $X = \{X^{\mathcal{M}}, X^{\mathcal{T}}, X^{\mathcal{L}}\}$, where $X^{\mathcal{M}}$, $X^{\mathcal{T}}$, and $X^{\mathcal{L}}$ correspond to sequences of metrics, traces, and logs, respectively. Let $\mathcal{V}$ denote the set of instances. Specifically, $X^{D}\in \mathbb{R}^{|\mathcal{V}|\times N^D\times T}$, where $|\mathcal{V}|$ is the total number of instances and $T$ is the number of snapshots. Note that $N^D$ is the modality-specific dimension of the modality $D$. For example, $N^\mathcal{L}$ denotes the number of log templates.

\subsubsection{Inter-Modality Feature Fusion}
This module is tasked with building the cross-modal relationship between multimodal telemetry. It comprises two parts: encoders for temporal feature extraction and an attention mechanism for multimodal fusion. Given sequences of three modalities $X = \{X^{\mathcal{M}}, X^{\mathcal{T}}, X^{\mathcal{L}}\}$, we define the snapshot feature at the time step $t$ ($t\in[1,T]$) as $X_t^D$, where $D \in \{\mathcal{M}, \mathcal{T}, \mathcal{L}\}$ and $X_t^D\in\mathbb{R}^{|\mathcal{V}|\times N^D}$. We employ gated recurrent units (GRUs) as the encoder, which can effectively capture sequential dependencies, handling the temporal feature of each modality $D$ in turn. This process can be formulated as:

\begin{equation}
    z^D_t = \sigma(W^D_z [h^D_{t-1}, X^D_t]),
\end{equation}
\begin{equation}
    r^D_t = \sigma(W^D_r [h^D_{t-1}, X^D_t]),
\end{equation}
\begin{equation}
    \tilde{h}^D_t = \tanh\left( W^D [r^D_t \odot h^D_{t-1}, X^D_t]\right),
\end{equation}
\begin{equation}
    h^D_t = (1 - z^D_t) \odot \tilde{h}^D_t + z^D_t \odot h^D_{t-1},
\end{equation}
where $\{W_z^D, W_r^D, W^D\}$ represent the learnable recurrent weight matrices. $\sigma(\cdot)$ and $\odot$ denote the logistic sigmoid function and the Hadamard product, respectively. $z_t^D$ and $r_t^D$ represent the update and reset gates. Note that $\tilde{h}_t^D$ is a candidate activation computed based on the current input and the previous hidden state modified by the reset gate. With the help of the GRU cell, we can encapsulate the temporal information within $X^D\in \mathbb{R}^{|\mathcal{V}| \times N^D \times T}$, yielding $h_t^D \in \mathbb{R}^{|\mathcal{V}| \times d}$, which represents the new hidden state at time step $t$. $d$ is the predefined dimension for feature $h_t^D$.



CCLH uses the hidden states of the final time step $T$, denoted $h_T^\mathcal{M}$, $h_T^\mathcal{T}$, and $h_T^\mathcal{L}$, as latent features that contain temporal information from this failure period. Rather than directly concatenating multimodal features, CCLH introduces an attention mechanism to automatically bridge the gap between heterogeneous modalities:

\begin{equation}
    \alpha^D = \sigma(W_ah^D_T)
\end{equation}
\begin{equation}
\label{eq:attention-soft}
    \hat\alpha^D = \frac{exp(\alpha^D)}{\sum_{D'\in{\{\mathcal{M},\mathcal{T},\mathcal{L}\}}}exp(\alpha^D)},
\end{equation}
\begin{equation}
    H = \sum_{D\in{\{\mathcal{M},\mathcal{T},\mathcal{L}\}}}\hat{\alpha}^Dh^D_T,
\end{equation}
where $W_a$ denotes a learnable shared attention matrix, which computes modality-specific attention scores $\alpha^D$ for the final hidden states $h_T^D$. Eq. \ref{eq:attention-soft} normalizes all attention scores through a softmax function. Finally, CCLH performs a weighted summation for all modality-specific features to get the fused feature $H$, where $H\in\mathbb{R}^{|\mathcal{V}|\times d}$ and the weights correspond to the normalized attention scores $\hat{\alpha}^D$.

\begin{figure}
    \centering
    \includegraphics[width=1\linewidth]{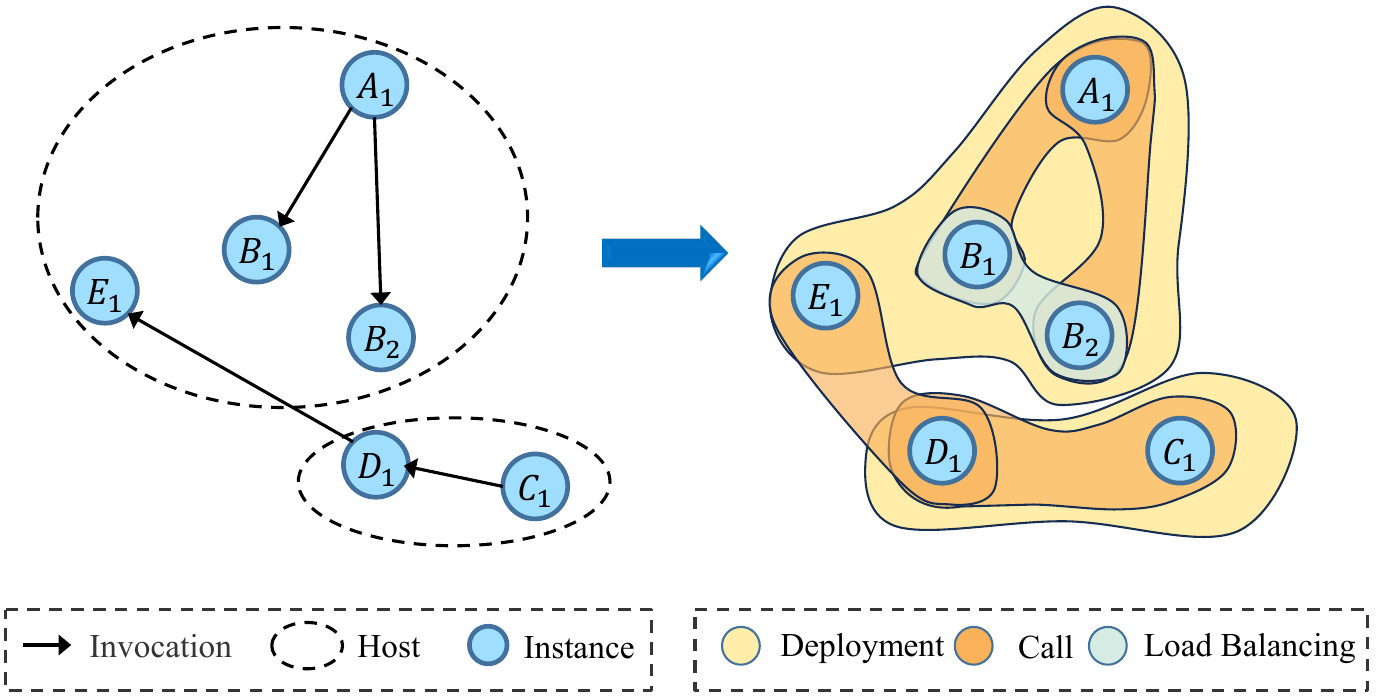}
    \caption{Example of hypergraph construction. The left part of this figure illustrates the dependency graph of one MSS, capturing partial relationships between instances. The right part shows the corresponding hypergraph transformed from this dependency graph.}
    \label{fig:hypergraph}
    \vspace{-4mm}
\end{figure}

\subsection{Hypergraph-based Status Fusion}
\label{sec:HSF}

\subsubsection{Hypergraph Construction} 
Inspired by \cref{sec:motivation2}, we propose a three-level taxonomy to characterize group relationships between instances, which manifests potential propagation paths of failures. As illustrated in Fig. \ref{fig:hypergraph}, the taxonomy categorizes inter-instance group relationships into three types: call, deployment, and load balancing. To represent and capture these relationships, CCLH extracts dependencies from traces and deployment configurations, thereby constructing a hypergraph $\mathcal{G}=(\mathcal{V}, \mathcal{E})$, where $\mathcal{V}$ denotes the set of instances, and $\mathcal{E}$ is the set of three types of hyperedge, defined as follows:

\begin{itemize}
    \item \textbf{Call}. The call is the most straightforward and intuitive relationship between instances. For example, as illustrated in Fig. \ref{fig:hypergraph}, CPU overhead in $E_1$ may lead to prolonged response times, thus degrading the performance of its upstream callers (e.g., $D_1$) through synchronous invocations. Accordingly, we represent such group influences by linking the instance and its associated callers through a call hyperedge.
    \item \textbf{Deployment}. As demonstrated in \cref{sec:motivation2}, co-located instances on the same host compete for shared resources such as CPU time, memory, and bandwidth, even though there may be no direct call relationship between them (e.g., $B_1$ and $E_1$). For all co-located instances that are deployed on the same host, we weave them with a deployment hyperedge to indicate the resource contention among them.
    \item \textbf{Load Balancing}. Despite the operational independence of each instance within a microservice, concealed relationships are inevitable due to the balanced nature of traffic routing strategies. For example, if $B_2$ encounters a pod failure, all incoming requests are routed to its sibling instance $B_1$. After a period of time, $B_1$ may experience degraded performance due to the overwhelming workload, potentially propagating failures to upstream components such as $A_1$ through call relationships. To handle this scenario, we group instances by their superordinate microservices based on information from the deployment, and introduce a hyperedge to model the sibling relationships. For example, a hyperedge is added between $B1$ and $B2$ to represent their shared affiliation with the same microservice, capturing their implicit relationships.
\end{itemize}

\subsubsection{Instance Status Fusion} 
Given the constructed hypergraph $\mathcal{G}=(\mathcal{V},\mathcal{E})$ with the fused feature $H$, we implement a heterogeneous hypergraph attention network, namely UniGAT-HE, aggregating spatio patterns from hyperedges and simulating failure backtracking. For the fused feature $H\in\mathbb{R}^{|\mathcal{V}|\times d}$, we assign each instance $v$ its corresponding embedding $h_v$ ($h_v\in\mathbb{R}^d$) along the first dimension. 

Unlike traditional graph neural networks that update node embeddings by directly merging information from neighborhoods, UniGAT-HE first aggregates information in each hyperedge. For hyperedge $e\in\mathcal{E}$, we calculate its own embedding $f_e$ by summarizing features of its associated instances:
\begin{equation}
    f_e = \frac{1}{|e|}\sum_{j\in e}h_{j},
\end{equation}
where $h_j$ is the embedding of instance $j$ and $|e|$ denotes the number of instances affiliated with this hyperedge. 

Although HGNNs \cite{ijcai21-UniGNN} effectively facilitate message passing in hypergraphs, they struggle to differentiate between different types of hyperedges, which typically share a unified attention weight in implementation. Therefore, traditional HGNNs can hardly represent our three-level taxonomy for group relationships between instances. To enable fine-grained modeling of heterogeneous hyperedges and capture diverse failure propagation patterns, UniGAT-HE introduces distinct attention mechanisms for each hyperedge type. Let $a_{t(e)}$ represent the attention weight of hyperedge $e$, where $t(e)$ denotes the type of $e$, including call, deployment, and load balancing. We update each instance embedding $h_v$:

\begin{equation}
    \alpha_{ve} = \sigma(a^T_{t(e)}[Wh_v || Wf_e]),
\end{equation}
\begin{equation}
    \hat{\alpha_{ve}} = \frac{exp(\alpha_{ve})}{\sum_{e'\in\mathcal{E}(v)}exp(\alpha_{ve'})},
\end{equation}
\begin{equation}
    h_v = \sum_{e\in\mathcal{E}(v)}\hat{\alpha_{ve}}Wf_e,
\end{equation}
where $[\cdot || \cdot]$ denotes the concatenation of two embeddings; $W$ represents the learnable parameters for linear transformation; and $\sigma$ denotes the nonlinear activation function. After computing the attention score $\alpha_{ve}$ between instance $v$ and hyperedge $e$, we subsequently apply the Softmax function to normalize the scores across all hyperedges ($\mathcal{E}(v)$) connected to $v$, yielding $\hat{\alpha_{ve}}$ that indicates the relative importance of hyperedge $e$ to instance $v$. Eventually, we update the instance embedding $h_v$ by assimilating features from all associated hyperedges in a weighted manner.

\subsection{Cascaded Conditional Learning and Diagnosis}
\label{sec:CCL}

Based on the set of updated instance embeddings $H=\{h_v|v\in\mathcal{V}\}$, the subsequent action is to perform root cause localization (RCL) and failure type identification (FTI) at the instance level. To achieve this, CCLH incorporates a scorer for RCL and a classifier for FTI using two multilayer perceptrons (MLPs). Motivated by \cref{sec:motivation1}, we perform joint learning for these two tasks and orchestrate their diagnostic stages in a sequential manner, facilitating not only the utilization of shared knowledge but also alignment with the practical diagnostic workflow. 

During the training phase, CCLH first optimizes the scorer separately and freezes the classifier. Considering the dynamic nature of instances in MSS, we set the output size of the scorer to 1, rather than prematurely fixing it to the number of instances. The input size of the scorer is set to the dimension of $h_v$. Consequently, we feed each instance embedding $h_v$ into the scorer to predict a suspect score $s_v$, which reflects the likelihood that the instance is the culprit component. The RCL task is inherently a multiclass classification problem, as it involves collecting a set of scores $\mathcal{S} = \{s_1, s_2, \dots, s_v|v \in \mathcal{V} \}$ for all instance embeddings and then ranking them to identify the most likely culprit. Hence, we adopt the cross-entropy loss to optimize the deviation between the ground-truth and the predicted culprit component:

\begin{equation}
    \mathcal{L}_1=-\frac{1}{N}\sum_{i=1}^{N}\sum_{j=1}^{|\mathcal{V}|}y_{j}^ilog(p_{j}^i),
\end{equation}
where $N$ is the total number of failure cases in the training dataset; $y_i^{j}$ equals 1 if instance $j$ is the culprit component in the $i$-th failure case, and 0 otherwise; $p_i^{j}$ denotes the predicted probability of instance $j$ being the culprit component.

CCLH continuously optimizes the scorer until its performance, measured by $HR@1$ as described in \cref{sec:exp}, exceeds the predefined condition that is quantified by a task trigger $\theta$. $\theta$ is a hyperparameter that can be adjusted manually. Note that a higher $\theta$ in the training set does not necessarily indicate better final performance, as we must consider generalizability and avoid overfitting. After that, we introduce the optimization of the classifier for FTI. Unlike the scorer that scores each instance, the classifier receives the embedding of the culprit component and directly outputs the probabilities of each failure type. We also employ cross-entropy loss to guide the optimization for the classifier and update the total loss as:

\begin{equation}
    \mathcal{L}=\mathcal{L}_1+ (-\frac{1}{N}\sum_{i=1}^{N}\sum_{k=1}^{|\mathcal{V}|}y_{k}^ilog(p_{k}^i)),
\end{equation}
where $p_i^{k}$ denotes the predicted probability of type $k$ being the real failure type; $y_i^{k}$ equals 1 if type $k$ is the failure type in the $i$-th failure case, and 0 otherwise. During the inference phase, we also first locate the culprit component using the scorer and then utilize the classifier to identify its failure type.

\begin{table*}[t]
  \centering
  \caption{Performance comparison across baselines.}
  \begin{threeparttable}
  \resizebox{\linewidth}{!}{
    \begin{tabular}{c|cccccc|cccccc|cccccc}
    \toprule
    \multirow{2}[4]{*}{Approach} & \multicolumn{6}{c|}{$\mathcal{A}$}                        & \multicolumn{6}{c|}{$\mathcal{B}$}                        & \multicolumn{6}{c}{$\mathcal{C}$} \\
\cmidrule{2-19}          & HR@1  & HR@3  & Avg@3 & Pre   & Rec   & F1    & HR@1  & HR@3  & Avg@3 & Pre   & Rec   & F1    & HR@1  & HR@3  & Avg@3 & Pre   & Rec   & F1 \\
    \midrule
    MicroRCA & 0.207  & 0.449  & 0.328  & -     & -     & -     & 0.061  & 0.231  & 0.146  & -     & -     & -     & 0.050  & 0.154  & 0.101  & -     & -     & - \\
    LogCluster & -     & -     & -     & 0.485  & 0.503  & 0.485  & -     & -     & -     & 0.242  & 0.246  & 0.243  & -     & -     & -     & 0.167 & 0.168 & 0.167 \\
    DiagFusion & 0.465  & 0.886  & 0.726  & 0.937  & 0.934  & 0.933  & 0.533  & 0.718 & 0.625  & 0.479  & 0.456  & 0.464  & 0.398  & 0.631  & 0.522  & 0.379 & 0.358 & 0.364 \\
    TVDiag & 0.811  & 0.939  & 0.884  & \textbf{0.976} & \textbf{0.977} & \textbf{0.977} & 0.834  & 0.925  & 0.884  & 0.585  & 0.568  & 0.577  & 0.832  & 0.904  & 0.877  & 0.510  & 0.500   & 0.505 \\
    Medicine & -     & -     & -     & 0.810  & 0.806  & 0.805  & -     & -     & -     & 0.492  & 0.492  & 0.486  & -     & -     & -     & 0.486 & 0.480  & 0.463 \\
    DeepHunt & 0.308      & 0.554      &  0.448     & -     & -     & -     & 0.436      &  0.708     &  0.588     & -     & -     & -     &  0.441     &   0.617    &  0.593    & -     & -     & - \\
    CCLH  & \textbf{0.875} & \textbf{0.950} & \textbf{0.920} & 0.944  & 0.941  & 0.941  & \textbf{0.923} & \textbf{0.954} & \textbf{0.938} & \textbf{0.777} & \textbf{0.764} & \textbf{0.768} & \textbf{0.918} & \textbf{0.950} & \textbf{0.938} & \textbf{0.778} & \textbf{0.774} & \textbf{0.772} \\
    \bottomrule
    \end{tabular}}%
    \begin{tablenotes}
    \footnotesize
    \item \textit{Note:} ``-'' represents that there is no value in the corresponding cell. Boldface highlights the best performance for each metric.
\end{tablenotes}
	\end{threeparttable}
  \label{tab:performance}%
  \vspace{-4mm}
\end{table*}%

\section{Evaluation}
\label{sec:exp}
In this section, we evaluate the effectiveness of CCLH by answering five research questions (RQs):

\begin{enumerate}
    \item RQ1: How does CCLH perform on two diagnostic tasks?

    \item RQ2: What is the effectiveness of each component in CCLH?

    \item RQ3: How does the task trigger that governs cascaded task learning affect the overall performance?

     \item RQ4: Can CCLH generalize to scenarios involving previously unseen culprit components during inference?

     \item RQ5: How does CCLH perform in terms of efficiency?
\end{enumerate}

\subsection{Experimental Design}
\textbf{Dataset}. We conducted extensive experiments on one public dataset and two datasets collected from popular benchmarks.

\begin{itemize}
    \item Dataset $\mathcal{A}$. This dataset\footnote{https://github.com/CloudWise-OpenSource/GAIA-DataSet} contains 1,099 labeled failure cases and the corresponding multimodal telemetry for two weeks, acquired from an online system that includes 10 instances and 5 failure types. 
    \item Datasets $\mathcal{B}$ and $\mathcal{C}$. They were collected from two open source benchmarks, namely Online Boutique \cite{hipster} and SockShop\footnote{https://github.com/microservices-demo/microservices-demo}. Following previous work \cite{yu2023nezha, xie2024tvdiag, lee2023eadro, pham2024baro}, we separately deployed these two MSSs in our cluster and deliberately injected failures based on Chaos Mesh\footnote{https://chaos-mesh.org}. Failures involve typical resource overhead (CPU, memory), network issues (loss, delay, and corruption), and pod/container faults. We injected failures into each instance and repeated five times, resulting in two datasets comprised of 489 and 700 failure cases, respectively.
\end{itemize}

\textbf{Baseline Methods}. We selected several state-of-the-art diagnostic methods as baselines, including two single-modal approaches and four multimodal methods for comparison.

\begin{itemize}
    \item MicroRCA \cite{wu2020microrca}: It is a metric-based approach that incorporates the Personalized PageRank algorithm into RCL.
    \item LogCluster \cite{lin2016log}: It annotates historical failures and clusters the corresponding logs, establishing a knowledge base capable of identifying failure types through log matching.
    \item DiagFusion \cite{zhang2023robust}: It is a multimodal method for RCL and FTI, which captures events in failure periods and adopts two independent topology adaptive graphs for feature learning.
    \item TVDiag \cite{xie2024tvdiag}: It presents a task-oriented and view-invariant feature learning method for RCL and FTI, capturing the intermodal and task-model relationships.
    \item Medicine \cite{tao2024giving}: It is an FTI method that balances the bias of modalities based on multimodal adaptive optimization.
    \item DeepHunt \cite{sun2025interpretable}: It is a multimodal RCL method that scores instances based on reconstruction errors and modeled failure propagation. To ensure fairness, we adopt the supervised variant of Deephunt.
\end{itemize}

\textbf{Evaluation Metrics}. Since FTI is inherently a multiclass classification task, we adopted standard classification metrics for evaluation: $Pre=\frac{TP}{TP+FP}$ and $Rec=\frac{TP}{TP+FN}$, where $TP$ denotes the number of true positives and $FP$ represents the number of false positives. To align with the preceding work \cite{xie2024tvdiag, zhang2023robust}, we used a weighted average F1-score to account for class imbalance and provide a more comprehensive evaluation. 

Regarding RCL, we utilized the hit ratio $HR@k=\frac{1}{N}\sum_{i=1}^{N}r_i\in \mathcal{R}_i[1:k]$, where $N$ is the total number of failure cases, $r_i$ is the label of the $i$-th faliure case, and $\mathcal{R}_i[1:k]$ is the top k instances in the sorted results of scores $\mathcal{S}$. We further introduced $Avg@k=\frac{1}{k}\sum_{i=1}^{k}HR@i$ as the average performance of the hit ratio.

\textbf{Implementation Details}.
We implemented CCLH using Python 3.8, PyTorch 1.12.1, and DGL 0.9.1. All experiments were conducted on a Windows laptop equipped with a NVIDIA GeForce RTX 2070 GPU, a 16-core Intel(R) Core(TM) i7-10875H CPU, and 16 GB RAM. The GRU module was configured with 3 layers and a hidden dimension of 256. For UniGAT-HE, we used a 2-layer architecture with a hidden dimension of 256. To accelerate training convergence, we applied an early stopping strategy, terminating training when the loss plateaued on the training set. Each dataset was split into 60\% for training and 40\% for testing.

\subsection{RQ1: Performance Comparison \& Ablation Study}
\label{sec:RQ1}
In this section, we evaluated selected approaches on three datasets. Table \ref{tab:performance} summarizes the performance of different baselines. 

In terms of RCL, multimodal methods consistently outperform single-modal methods in both $HR@k$ and $Avg@3$, owing to the ability to incorporate information from multiple views. CCLH yields a higher hit ratio than other baselines across all datasets, which is attributed to its granular taxonomy for inter-instance relationships that can effectively capture propagation patterns of failures. Moreover, both CCLH and TVDiag demonstrate superior localization accuracy compared to other multimodal baselines, as they leverage instance-level features to pinpoint the culprit component. In contrast, DiagFusion relies on graph-level features aggregated via a pooling layer over all instance-level features, leading to information loss and degraded performance when faced with a large number of instances in MSS.

With respect to FTI, CCLH significantly surpasses baselines in almost all datasets. In the case of dataset $\mathcal{A}$, CCLH performs comparably to other competitors, because the limited number of instances (only 10) in this MSS renders the failure scenario relatively less diverse. Notably, TVdiag, Medicine, and DiagFusion perform poorly on datasets $\mathcal{B}$ and $\mathcal{C}$, which involve more complex instance interactions. This is primarily because they perform FTI in parallel with RCL, requiring failure types to be inferred from graph-level features, which may dilute instance-specific signals. Compared to these methods, CCLH exhibits a significant improvement, achieving at least a 19.1\% increase in F1-score. The results align with our insight in \cref{sec:motivation2} that performing RCL first can effectively narrow the diagnostic scope, allowing FTI to focus on relevant instance-level information. 

\begin{figure*}[t]
    \centering
    \includegraphics[width=1\linewidth]{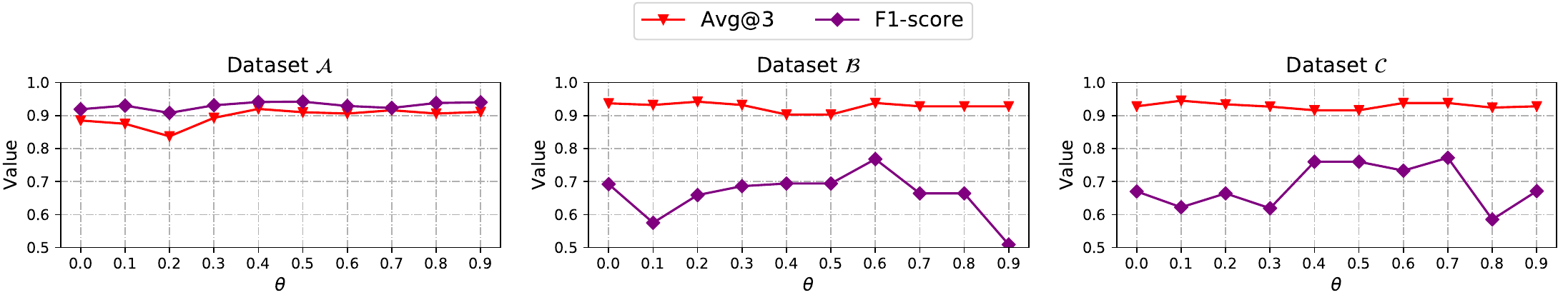}
    \caption{Performance comparison across different task triggers.}
    \label{fig:taskTrigger}
\end{figure*}

\begin{table}[t]
  \centering
  \caption{Ablation Study.}
  \begin{threeparttable}
  \resizebox{\linewidth}{!}{
    \begin{tabular}{c|cc|cc|cc}
    \toprule
    \multirow{2}[4]{*}{Approach} & \multicolumn{2}{c|}{$\mathcal{A}$} & \multicolumn{2}{c|}{$\mathcal{B}$} & \multicolumn{2}{c}{$\mathcal{C}$} \\
\cmidrule{2-7}          & Avg@3 & F1    & Avg@3 & F1    & Avg@3 & F1 \\
    \midrule
    CCLH $w/o$ CH & 0.875  & 0.933  & 0.921  & 0.633  & 0.930  & 0.728  \\
    CCLH $w/o$ DH & \textbf{0.923} & 0.935  & 0.932  & 0.679  & 0.932  & 0.687  \\
    CCLH $w/o$ LH & 0.841  & 0.898  & 0.911  & 0.673  & 0.904  & 0.719  \\
    CCLH $w/o$ HE & 0.905  & 0.935  & 0.915  & 0.709  & 0.888  & 0.708  \\
    CCLH $w/o$ HG & 0.838  & 0.860  & 0.800  & 0.449  & 0.863  & 0.555  \\
    CCLH $w/o$ TT & 0.885  & 0.919  & 0.937  & 0.692  & 0.928  & 0.670  \\
    CCLH  & 0.920  & \textbf{0.941} & \textbf{0.938} & \textbf{0.768} & \textbf{0.938} & \textbf{0.772} \\
    \bottomrule
    \end{tabular}}%
  \label{tab:ablation}%
    \begin{tablenotes}
    \footnotesize
    \item \textit{Note:} Boldface highlights the best performance for each metric.
\end{tablenotes}
  \end{threeparttable}
  \vspace{-2mm}
\end{table}%

\subsection{RQ2: Ablation Study}

To demonstrate the effects of key components of CCLH, we further conducted extensive ablation studies across three datasets. The variants are described as follows:

\begin{itemize}
    \item CCLH $w/o$ CH: Removes the call hyperedges from the hypergraph.
    \item CCLH $w/o$ DH: Removes the deployment hyperedges in the hypergraph.
    \item CCLH $w/o$ LH: Removes the load balancing hyperedges in the hypergraph.
    \item CCLH $w/o$ HG: Replaces the hypergraph with a directed graph and uses GAT as the backbone.
    \item CCLH $w/o$ HE: Substitutes the UniGAT-HE module with a standard UniGAT for hypergraph learning.
    \item CCLH $w/o$ TT: Removes the task trigger $\theta$ and learns two tasks simultaneously.
\end{itemize}

The ablation results are shown in Table \ref{tab:ablation}. We select $Avg@k$ and F1-score as the measurements of the variants. It can be observed that all components contribute positively to the final performance, demonstrating their effects on CCLH.

To assess the proposed hyperedges defined in \cref{sec:HSF}, we ablated them individually and further replaced them with common directed edges. The results show that most variants exhibit a performance decrease when hyperedges are removed. In particular, variant $w/o$ LH exhibits a substantial performance decline across all datasets, underscoring the importance of modeling the influence of sibling instances in hypergraph construction. The variant $w/o$ HG exhibits a macroscopic performance degradation, with an average decrease of 9.8\% in $Avg@3$ and 20.6\% in F1-score, which manifests the superiority of modeling group relationships over point-to-point relationships, further demonstrating the insight proposed by \cref{sec:motivation1}.

In addition, we measured the effectiveness of the proposed UniGAT-HE and the task trigger $\theta$. CCLH achieves superior results compared to variant $w/o$ HE. This is predominantly because our UniGAT-HE can effectively capture high-order hyperedge potentiality through an attention mechanism, thereby endowing CCLH with the ability to distinguish hyperedge weights across diverse failure scenarios. Section \ref{sec:motivation2} argues that diagnostic tasks should be organized in alignment with their causal order. This insight is validated by the performance of the variant $w/o$ TT, which suffers due to premature involvement of FTI during training. The underlying rationale for deferring the training of FTI is that CCLH relies on instance-level features from the suspected culprit component, which are identified by the RCL model. In the early training stages, the RCL model is still under-fitting, resulting in low-confidence localization and unreliable inputs for FTI.

\begin{figure*}[t]
    \centering
    \includegraphics[width=1\linewidth]{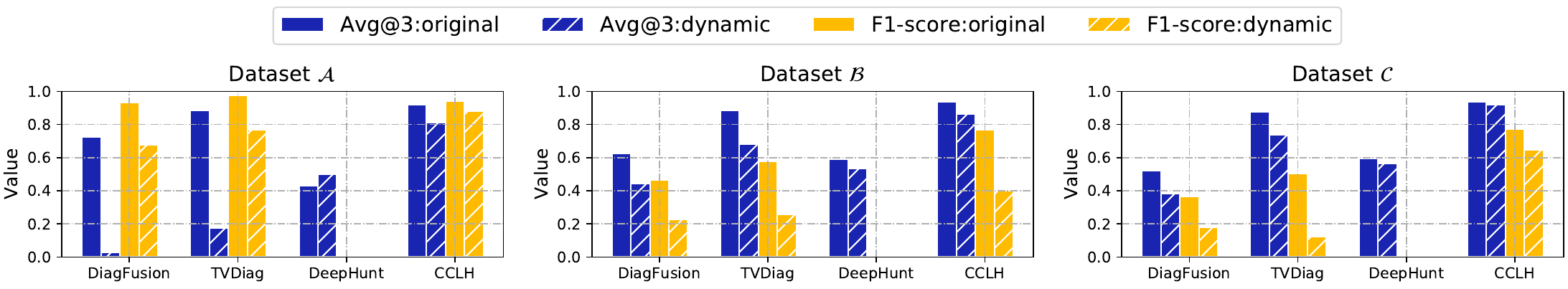}
    \caption{Performance variation when encountering unseen culprit components. ``original'' and ``dynamic'' indicate the performance under the original and resplit (unseen component) settings, respectively.}
    \label{fig:dynamic}
    \vspace{-3mm}
\end{figure*}

\subsection{RQ3: Sensitivity to Task Trigger}
Task trigger $\theta$ serves as a crucial control knob that governs when the FTI classifier participates in the cascaded conditional learning procedure. Fig. \ref{fig:taskTrigger} shows the impact of different $\theta$ on the performance of CCLH. We can deduce that the RCL module is insensitive to the $\theta$ setting, as it has already been optimized independently before meeting this threshold. In contrast, the performance of FTI exhibits more pronounced fluctuations across different $\theta$ values, highlighting its reliance on the maturity of the RCL model. We find that a higher $\theta$ does not necessarily yield better FTI performance. This is mainly because prolonged independent optimization of RCL can lead to overfitting, reducing the generalizability of both the scorer and the classifier. Nonetheless, the overall performance of CCLH still remains steady across different $\theta$ settings, indicating the robustness and stability of CCLH. According to Fig. \ref{fig:taskTrigger}, we set $\theta$ as 0.4 for dataset $\mathcal{A}$, 0.6 for datset $\mathcal{B}$, and 0.7 for dataset $\mathcal{C}$, respectively. Developing an automatic method for selecting $\theta$ to minimize manual tuning remains an important direction for future work.

\subsection{RQ4: Generalization Capability} 
Although we can simulate certain failures using chaos engineering techniques, unforeseen failures can still occur in production environments. For example, a known failure type arises in a microservice that was previously unaffected. To evaluate the generalization capability of CCLH in such scenarios involving unseen culprit components, we resplit all datasets accordingly. Specifically, for each failure type, we randomly select 60\% of the culprit components for training, while reserving the remaining 40\% for testing. This setting simulates cases where certain culprit components are not encountered during training.

Fig. \ref{fig:dynamic} illustrates the performance comparison between the original and reorganized datasets. DiagFusion and TVDiag exhibit a significant decline in two metrics across all datasets. This is primarily attributed to their alert embedding mechanism, which fails to generalize to new alerts absent from the training datasets. In contrast, CCLH and DeepHunt achieve more flexible adaptation to new patterns and ensure more stable performance, owing to their direct transformation of multimodal telemetry into learnable embeddings, facilitating more flexible adaptation to new patterns and ensuring more stable performance. It is worth noting that FTI performance shows a more pronounced decline compared to RCL in CCLH, since features from different culprit components inevitably exhibit tremendous discrepancies that hinder generalization. In summary, CCLH demonstrates strong generalizability when faced with previously unseen culprit components.

\begin{table}[t]
  \centering
  \caption{Statistics about the training and inference time consumption of deep learning-based baselines.}
  \resizebox{\linewidth}{!}{
    \begin{tabular}{c|ccc}
    \toprule
    \multicolumn{1}{c|}{\multirow{2}[4]{*}{Approach}} & \multicolumn{3}{c}{Dataset $\mathcal{A}$} \\
\cmidrule{2-4}    \multicolumn{1}{c|}{} & Offline training time (s) & Training per epoch (s) & Inference per case (s) \\
    \midrule
    DiagFusion & 19.687  & 0.030  & 0.013  \\
    TVDiag &   386.176    &  2.645     & 0.408 \\
    Medicine & 7987.860  & 26.626  & 0.387  \\
    DeepHunt & 685.462  & 3.808  & 0.505  \\
    CCLH  & 168.204  & 1.665  & 0.434  \\
    \bottomrule
    \end{tabular}}%
  \label{tab:efficiency}%
  \vspace{-4mm}
\end{table}%

\subsection{RQ5: Efficiency}
Table \ref{tab:efficiency} summarizes the time consumption of the training and inference phases in deep learning-based diagnostic models. We only present the results of dataset $\mathcal{A}$ because there is little divergence between these datasets.

In terms of training cost per epoch, DiagFusion achieves the quickest iteration speed due to its lightweight network structure, which comprises only two topology adaptive graphs and two MLPs. The training time of Medicine is longer than that of others, because it involves a gradient suppression module to adjust learning rates based on evaluation after each epoch, which introduces additional computational overhead, but remains within an acceptable range. CCLH exhibits modest efficiency because it locates culprit components by sorting scores of instances, achieving the trade-off between dynamicity and efficiency. Nevertheless, the total offline training time remains efficient, as CCLH leverages an early stopping strategy that terminates training once model convergence is detected. Due to the infrequent nature of model updates, the time consumption of the training phase is acceptable.

High inference costs are prohibitive, especially in production environments with real-time requirements. Generally, all models can diagnose failures within one second due to their lightweight network structures, which are sufficiently efficient in RCA.

\section{Discussion}

\subsection{limitation}
There are two primary limitations in CCLH: (1) Integration for platform events. CCLH perceives the status of the MSS through metrics, traces, and logs, focusing primarily on application-side performance. However, certain failures caused by software changes are more apparent on the platform side \cite{huang2024faasrca}. For example, SREs can often identify root causes by checking the corresponding change events on the platform \cite{yu2024changerca}. This presents a limitation due to the absence of platform events in our datasets. In production environments, CCLH can mitigate this limitation by incorporating such events through natural language processing techniques, as demonstrated in previous studies \cite{zhang2023robust, xie2024tvdiag}. (2) Joint evaluation of diagnostic tasks. Despite comprehensive assessments of individual tasks, there are no established indicators that measure the joint performance of both tasks. In future work, we plan to investigate how to effectively integrate the outcomes of both tasks and assess the step-wise cost along the actual diagnostic trajectory.

\subsection{Threats to Validity}
The internal threat lies primarily in the task trigger $\theta$, which governs the activation of diagnostic tasks. An inappropriate setting of $\theta$ can degrade the performance of CCLH. For example, an extremely high $\theta$ may prevent the subsequent FTI task from being triggered. To alleviate this threat, we recommend that the SREs perform a top-down search for suitable values of $\theta$, gradually decreasing it until the RCL task achieves the required efficacy, thus triggering the FTI task. 

The external threat mainly concerns the generalizability and practicality of CCLH in real-world MSS. For generalizability, we utilize one public dataset and two datasets collected from online benchmarks, which, however, may not fully represent the diversity of software systems in real-world scenarios. This limitation can be alleviated by injected failures, which are representative of common issues in industrial MSS and are widely adopted in chaos engineering practices. In practical settings, CCLH often operates under a few-shot or even zero-shot scenario, where there is a complete lack of annotated failure data. To alleviate this threat, SREs can proactively collect and annotate training data from the testing environment using chaos engineering techniques.

\section{Related Work}
Recently, significant efforts have been devoted to automated root cause analysis, which encompasses two critical tasks: root cause localization (RCL) \cite{han2024holistic,lee2023eadro,wang2024mrca,somashekar2024gamma,sun2024art,wang2020root,zheng2024mulan,yao2024chain,zhu2024hemirca,xie2024microservice,li2022actionable,tao2024diagnosing} and failure type identification (FTI) \cite{lin2016log,yuan2019approach,sui2023logkg}.

For the FTI task, we categorize the relevant work into two types: heuristic methods and deep learning-based methods. (1) Heuristic methods. MicroCBR \cite{liu2022microcbr} constructs a spatio-temporal knowledge graph of failures and uses fault fingerprints to retrieve the most relevant historical cases for FTI. Nevertheless, the knowledge base adopted in this method is prone to obsolescence as microservice systems (MSS) continue to evolve. TrinityRCL \cite{gu2023trinityrcl} constructs a multilevel causal graph and incorporates a PageRank algorithm to localize root causes across hierarchical layers. Despite the effectiveness of this architecture, traditional graph traversal techniques often suffer from computational overhead in large-scale microservices. (2) Deep learning-based methods. Medicine \cite{tao2024giving} is a multimodal method for FTI that adaptively balances the learning rates of multimodal telemetry (metrics, logs, and traces) based on gradient suppression, allowing thorough exploration of the diagnostic potential of each modality. However, this adjustment mechanism is computationally sensitive, which inevitably increases the training overhead.

Regarding the RCL task, Nezha \cite{yu2023nezha} detects code-level culprits by identifying changes in event patterns. In addition, DeepHunt\cite{sun2025interpretable} is a self-supervised method that uses a graph autoencoder to learn a representation of normal data and identifies culprit pods through deviation detection. Although these methods are competitive in automated RCL, they solely regard the inter-instance relationship as a pairwise paradigm, ignoring the group attributes of instance-wise influences.


Recent studies have focused on learning RCL and FTI jointly. DiagFusion \cite{zhang2023robust} extracts events from multimodal telemetry for a unified representation. Subsequently, it incorporates two topology adaptive graphs to handle RCL and FTI, respectively. TVDiag \cite{xie2024tvdiag} correlates two diagnostic tasks and multimodal telemetry using contrastive learning, enabling an effective extraction of shared knowledge. TraFaultDia \cite{wang2025cross} introduces a meta-learning mechanism to tackle cross-system problems in root cause analysis, providing end-to-end RCL and FTI. OpenRCA \cite{xuopenrca} explores large language models (LLMs) for automated fault diagnosis. However, experimental results demonstrate suboptimal performance and high latency. Despite their successful combination for two tasks, they overlook the inherent causal order between RCL and FTI. Particularly, the parallel diagnostic setup compels the FTI model to assimilate information in a coarse-grained manner from all possible instances, resulting in noticeable performance degradation in scenarios with numerous unseen instances.

\section{Conclusion}
This paper proposes CCLH, a root cause analysis framework that orchestrates diagnostic tasks based on cascaded conditional learning with hypergraphs. CCLH incorporates a GRU-based feature extractor and cross-modal attention mechanism, distilling and fusing temporal information from multimodal telemetry data. We develop a three-level taxonomy for group relationships among instances and construct a hypergraph to model intricate dependencies. We further implement a heterogeneous hypergraph attention network that aggregates spatial features from diverse hyperedges. By sequentially optimizing RCL and FTI in causal order, CCLH enables FTI to benefit from the results of the previously optimized RCL. Experimental results in three datasets demonstrate the effectiveness of CCLH. In the future, we will explore how to integrate more modalities (e.g., platform events) and investigate the joint measurements for two tasks.

\bibliographystyle{IEEEtran}
\bibliography{CCLH}

@article{sun2025interpretable,
  title={Interpretable failure localization for microservice systems based on graph autoencoder},
  author={Sun, Yongqian and Lin, Zihan and Shi, Binpeng and Zhang, Shenglin and Ma, Shiyu and Jin, Pengxiang and Zhong, Zhenyu and Pan, Lemeng and Guo, Yicheng and Pei, Dan},
  journal={ACM Transactions on Software Engineering and Methodology},
  volume={34},
  number={2},
  pages={1--28},
  year={2025},
  publisher={ACM New York, NY}
}

@inproceedings{yu2023nezha,
  title={Nezha: Interpretable fine-grained root causes analysis for microservices on multi-modal observability data},
  author={Yu, Guangba and Chen, Pengfei and Li, Yufeng and Chen, Hongyang and Li, Xiaoyun and Zheng, Zibin},
  booktitle={Proceedings of the 31st ACM Joint European Software Engineering Conference and Symposium on the Foundations of Software Engineering},
  pages={553--565},
  year={2023}
}

@misc{github,
author={Scott Sanders},
title = {GitHub Availability Report: October 2021},
year=2025,
howpublished  = {\url{https://github.blog/news-insights/company-news/github-availability-report-october-2021}}
}

@article{zhang2024failure,
  title={Failure diagnosis in microservice systems: A comprehensive survey and analysis},
  author={Zhang, Shenglin and Xia, Sibo and Fan, Wenzhao and Shi, Binpeng and Xiong, Xiao and Zhong, Zhenyu and Ma, Minghua and Sun, Yongqian and Pei, Dan},
  journal={ACM Transactions on Software Engineering and Methodology},
  year={2024},
  publisher={ACM New York, NY}
}

@article{xie2024tvdiag,
  title={TVDiag: A Task-oriented and View-invariant Failure Diagnosis Framework with Multimodal Data},
  author={Xie, Shuaiyu and Wang, Jian and He, Hanbin and Wang, Zhihao and Zhao, Yuqi and Zhang, Neng and Li, Bing},
  journal={ACM Transactions on Software Engineering and Methodology},
  year={2025}
}

@article{zhang2023robust,
  title={Robust failure diagnosis of microservice system through multimodal data},
  author={Zhang, Shenglin and Jin, Pengxiang and Lin, Zihan and Sun, Yongqian and Zhang, Bicheng and Xia, Sibo and Li, Zhengdan and Zhong, Zhenyu and Ma, Minghua and Jin, Wa and others},
  journal={IEEE Transactions on Services Computing},
  volume={16},
  number={6},
  pages={3851--3864},
  year={2023},
  publisher={IEEE}
}

@inproceedings{sun2024art,
  title={ART: A Unified Unsupervised Framework for Incident Management in Microservice Systems},
  author={Sun, Yongqian and Shi, Binpeng and Mao, Mingyu and Ma, Minghua and Xia, Sibo and Zhang, Shenglin and Pei, Dan},
  booktitle={Proceedings of the 39th IEEE/ACM International Conference on Automated Software Engineering},
  pages={1183--1194},
  year={2024}
}

@inproceedings{tao2024giving,
  title={Giving Every Modality a Voice in Microservice Failure Diagnosis via Multimodal Adaptive Optimization},
  author={Tao, Lei and Zhang, Shenglin and Jia, Zedong and Sun, Jinrui and Ma, Minghua and Li, Zhengdan and Sun, Yongqian and Yang, Canqun and Zhang, Yuzhi and Pei, Dan},
  booktitle={Proceedings of the 39th IEEE/ACM International Conference on Automated Software Engineering},
  pages={1107--1119},
  year={2024}
}

@inproceedings{lee2023eadro,
  title={Eadro: An end-to-end troubleshooting framework for microservices on multi-source data},
  author={Lee, Cheryl and Yang, Tianyi and Chen, Zhuangbin and Su, Yuxin and Lyu, Michael R},
  booktitle={2023 IEEE/ACM 45th International Conference on Software Engineering (ICSE)},
  pages={1750--1762},
  year={2023},
  organization={IEEE}
}

@article{pham2024baro,
  title={Baro: Robust root cause analysis for microservices via multivariate bayesian online change point detection},
  author={Pham, Luan and Ha, Huong and Zhang, Hongyu},
  journal={Proceedings of the ACM on Software Engineering},
  volume={1},
  number={FSE},
  pages={2214--2237},
  year={2024},
  publisher={ACM New York, NY, USA}
}

@article{tao2024diagnosing,
  title={Diagnosing performance issues for large-scale microservice systems with heterogeneous graph},
  author={Tao, Lei and Lu, Xianglin and Zhang, Shenglin and Luan, Jiaqi and Li, Yingke and Li, Mingjie and Li, Zeyan and Yu, Qingyang and Xie, Hucheng and Xu, Ruijie and others},
  journal={IEEE Transactions on Services Computing},
  year={2024},
  publisher={IEEE}
}

@article{sui2023logkg,
  title={Logkg: Log failure diagnosis through knowledge graph},
  author={Sui, Yicheng and Zhang, Yuzhe and Sun, Jianjun and Xu, Ting and Zhang, Shenglin and Li, Zhengdan and Sun, Yongqian and Guo, Fangrui and Shen, Junyu and Zhang, Yuzhi and others},
  journal={IEEE Transactions on Services Computing},
  volume={16},
  number={5},
  pages={3493--3507},
  year={2023},
  publisher={IEEE}
}

@inproceedings{roy2024exploring,
  title={Exploring llm-based agents for root cause analysis},
  author={Roy, Devjeet and Zhang, Xuchao and Bhave, Rashi and Bansal, Chetan and Las-Casas, Pedro and Fonseca, Rodrigo and Rajmohan, Saravan},
  booktitle={Companion Proceedings of the 32nd ACM International Conference on the Foundations of Software Engineering},
  pages={208--219},
  year={2024}
}

@inproceedings{feng2019hypergraph,
  title={Hypergraph neural networks},
  author={Feng, Yifan and You, Haoxuan and Zhang, Zizhao and Ji, Rongrong and Gao, Yue},
  booktitle={Proceedings of the AAAI conference on artificial intelligence},
  volume={33},
  number={01},
  pages={3558--3565},
  year={2019}
}

@article{bai2021hypergraph,
  title={Hypergraph convolution and hypergraph attention},
  author={Bai, Song and Zhang, Feihu and Torr, Philip HS},
  journal={Pattern Recognition},
  volume={110},
  pages={107637},
  year={2021},
  publisher={Elsevier}
}

@inproceedings{ijcai21-UniGNN,
  title     = {UniGNN: a Unified Framework for Graph and Hypergraph Neural Networks},
  author    = {Huang, Jing and Yang, Jie},
  booktitle = {Proceedings of the Thirtieth International Joint Conference on
               Artificial Intelligence, {IJCAI-21}},
  year      = {2021}
}

@article{arya2020hypersage,
  title={Hypersage: Generalizing inductive representation learning on hypergraphs},
  author={Arya, Devanshu and Gupta, Deepak K and Rudinac, Stevan and Worring, Marcel},
  journal={arXiv preprint arXiv:2010.04558},
  year={2020}
}

@article{fifty2021efficiently,
  title={Efficiently identifying task groupings for multi-task learning},
  author={Fifty, Chris and Amid, Ehsan and Zhao, Zhe and Yu, Tianhe and Anil, Rohan and Finn, Chelsea},
  journal={Advances in Neural Information Processing Systems},
  volume={34},
  pages={27503--27516},
  year={2021}
}

@article{zhang2021survey,
  title={A survey on multi-task learning},
  author={Zhang, Yu and Yang, Qiang},
  journal={IEEE transactions on knowledge and data engineering},
  volume={34},
  number={12},
  pages={5586--5609},
  year={2021},
  publisher={IEEE}
}

@inproceedings{wu2020microrca,
  title={Microrca: Root cause localization of performance issues in microservices},
  author={Wu, Li and Tordsson, Johan and Elmroth, Erik and Kao, Odej},
  booktitle={NOMS 2020-2020 IEEE/IFIP Network Operations and Management Symposium},
  pages={1--9},
  year={2020},
  organization={IEEE}
}

@inproceedings{lin2016log,
  title={Log clustering based problem identification for online service systems},
  author={Lin, Qingwei and Zhang, Hongyu and Lou, Jian-Guang and Zhang, Yu and Chen, Xuewei},
  booktitle={Proceedings of the 38th international conference on software engineering companion},
  pages={102--111},
  year={2016}
}

@inproceedings{he2017drain,
  title={Drain: An online log parsing approach with fixed depth tree},
  author={He, Pinjia and Zhu, Jieming and Zheng, Zibin and Lyu, Michael R},
  booktitle={2017 IEEE international conference on web services (ICWS)},
  pages={33--40},
  year={2017},
  organization={IEEE}
}

@inproceedings{huang2024faasrca,
  title={FaaSRCA: Full Lifecycle Root Cause Analysis for Serverless Applications},
  author={Huang, Jin and Chen, Pengfei and Yu, Guangba and Wang, Yilun and Huang, Haiyu and He, Zilong},
  booktitle={2024 IEEE 35th International Symposium on Software Reliability Engineering (ISSRE)},
  pages={415--426},
  year={2024},
  organization={IEEE}
}

@article{yu2024changerca,
  title={ChangeRCA: Finding Root Causes from Software Changes in Large Online Systems},
  author={Yu, Guangba and Chen, Pengfei and He, Zilong and Yan, Qiuyu and Luo, Yu and Li, Fangyuan and Zheng, Zibin},
  journal={Proceedings of the ACM on Software Engineering},
  volume={1},
  number={FSE},
  pages={24--46},
  year={2024},
  publisher={ACM New York, NY, USA}
}

@misc{hipster,
author={GoogleCloudPlatform},
title = {OnlineBoutique},
year=2025,
howpublished  = {\url{https://github.com/GoogleCloudPlatform/microservices-demo}}
}

@article{gu2023trinityrcl,
  title={Trinityrcl: Multi-granular and code-level root cause localization using multiple types of telemetry data in microservice systems},
  author={Gu, Shenghui and Rong, Guoping and Ren, Tian and Zhang, He and Shen, Haifeng and Yu, Yongda and Li, Xian and Ouyang, Jian and Chen, Chunan},
  journal={IEEE Transactions on Software Engineering},
  volume={49},
  number={5},
  pages={3071--3088},
  year={2023},
  publisher={IEEE}
}

@article{han2024holistic,
  title={Holistic Root Cause Analysis for Failures in Cloud-Native Systems Through Observability Data},
  author={Han, Yongqi and Du, Qingfeng and Huang, Ying and Li, Pengsheng and Shi, Xiaonan and Wu, Jiaqi and Fang, Pei and Tian, Fulong and He, Cheng},
  journal={IEEE Transactions on Services Computing},
  year={2024},
  publisher={IEEE}
}

@inproceedings{liu2022microcbr,
  title={Microcbr: Case-based reasoning on spatio-temporal fault knowledge graph for microservices troubleshooting},
  author={Liu, Fengrui and Wang, Yang and Li, Zhenyu and Ren, Rui and Guan, Hongtao and Yu, Xian and Chen, Xiaofan and Xie, Gaogang},
  booktitle={International Conference on Case-Based Reasoning},
  pages={224--239},
  year={2022},
  organization={Springer}
}

@misc{zhang2024survey,
      title={A Survey of AIOps for Failure Management in the Era of Large Language Models}, 
      author={Lingzhe Zhang and Tong Jia and Mengxi Jia and Yifan Wu and Aiwei Liu and Yong Yang and Zhonghai Wu and Xuming Hu and Philip S. Yu and Ying Li},
      year={2024},
      eprint={2406.11213},
      archivePrefix={arXiv},
      primaryClass={cs.SE},
      url={https://arxiv.org/abs/2406.11213}, 
}

@inproceedings{zhang2024trace,
  title={Trace-based multi-dimensional root cause localization of performance issues in microservice systems},
  author={Zhang, Chenxi and Dong, Zhen and Peng, Xin and Zhang, Bicheng and Chen, Miao},
  booktitle={Proceedings of the IEEE/ACM 46th International Conference on Software Engineering},
  pages={1--12},
  year={2024}
}

@inproceedings{wang2024mrca,
  title={MRCA: Metric-level Root Cause Analysis for Microservices via Multi-Modal Data},
  author={Wang, Yidan and Zhu, Zhouruixing and Fu, Qiuai and Ma, Yuchi and He, Pinjia},
  booktitle={Proceedings of the 39th IEEE/ACM International Conference on Automated Software Engineering},
  pages={1057--1068},
  year={2024}
}

@inproceedings{somashekar2024gamma,
  title={GAMMA: Graph Neural Network-Based Multi-Bottleneck Localization for Microservices Applications},
  author={Somashekar, Gagan and Dutt, Anurag and Adak, Mainak and Lorido Botran, Tania and Gandhi, Anshul},
  booktitle={Proceedings of the ACM Web Conference 2024},
  pages={3085--3095},
  year={2024}
}

@inproceedings{wang2020root,
  title={Root-cause metric location for microservice systems via log anomaly detection},
  author={Wang, Lingzhi and Zhao, Nengwen and Chen, Junjie and Li, Pinnong and Zhang, Wenchi and Sui, Kaixin},
  booktitle={2020 IEEE international conference on web services (ICWS)},
  pages={142--150},
  year={2020},
  organization={IEEE}
}

@inproceedings{zheng2024mulan,
  title={MULAN: multi-modal causal structure learning and root cause analysis for microservice systems},
  author={Zheng, Lecheng and Chen, Zhengzhang and He, Jingrui and Chen, Haifeng},
  booktitle={Proceedings of the ACM Web Conference 2024},
  pages={4107--4116},
  year={2024}
}

@inproceedings{yao2024chain,
  title={Chain-of-event: Interpretable root cause analysis for microservices through automatically learning weighted event causal graph},
  author={Yao, Zhenhe and Pei, Changhua and Chen, Wenxiao and Wang, Hanzhang and Su, Liangfei and Jiang, Huai and Xie, Zhe and Nie, Xiaohui and Pei, Dan},
  booktitle={Companion Proceedings of the 32nd ACM International Conference on the Foundations of Software Engineering},
  pages={50--61},
  year={2024}
}

@article{zhu2024hemirca,
  title={HeMiRCA: Fine-grained root cause analysis for microservices with heterogeneous data sources},
  author={Zhu, Zhouruixing and Lee, Cheryl and Tang, Xiaoying and He, Pinjia},
  journal={ACM Transactions on Software Engineering and Methodology},
  volume={33},
  number={8},
  pages={1--25},
  year={2024},
  publisher={ACM New York, NY}
}

@inproceedings{xie2024microservice,
  title={Microservice root cause analysis with limited observability through intervention recognition in the latent space},
  author={Xie, Zhe and Zhang, Shenglin and Geng, Yitong and Zhang, Yao and Ma, Minghua and Nie, Xiaohui and Yao, Zhenhe and Xu, Longlong and Sun, Yongqian and Li, Wentao and others},
  booktitle={Proceedings of the 30th ACM SIGKDD Conference on Knowledge Discovery and Data Mining},
  pages={6049--6060},
  year={2024}
}

@inproceedings{li2022actionable,
  title={Actionable and interpretable fault localization for recurring failures in online service systems},
  author={Li, Zeyan and Zhao, Nengwen and Li, Mingjie and Lu, Xianglin and Wang, Lixin and Chang, Dongdong and Nie, Xiaohui and Cao, Li and Zhang, Wenchi and Sui, Kaixin and others},
  booktitle={Proceedings of the 30th ACM Joint European Software Engineering Conference and Symposium on the Foundations of Software Engineering},
  pages={996--1008},
  year={2022}
}

@inproceedings{xuopenrca,
  title={OpenRCA: Can Large Language Models Locate the Root Cause of Software Failures?},
  author={Xu, Junjielong and Zhang, Qinan and Zhong, Zhiqing and He, Shilin and Zhang, Chaoyun and Lin, Qingwei and Pei, Dan and He, Pinjia and Zhang, Dongmei and Zhang, Qi},
  booktitle={The Thirteenth International Conference on Learning Representations}
}

@inproceedings{wang2025cross,
  title={Cross-System Software Log-based Anomaly Detection Using Meta-Learning},
  author={Wang, Yuqing and M{\"a}ntyl{\"a}, Mika V and Nyyss{\"o}l{\"a}, Jesse and Ping, Ke and Wang, Liqiang},
  booktitle={2025 IEEE International Conference on Software Analysis, Evolution and Reengineering (SANER)},
  pages={454--464},
  year={2025},
  organization={IEEE}
}

@inproceedings{yuan2019approach,
  title={An approach to cloud execution failure diagnosis based on exception logs in openstack},
  author={Yuan, Yue and Shi, Wenchang and Liang, Bin and Qin, Bo},
  booktitle={2019 IEEE 12th International Conference on Cloud Computing (CLOUD)},
  pages={124--131},
  year={2019},
  organization={IEEE}
}
\end{document}